\documentclass[
twocolumn,
]{ceurart}

\usepackage{listings}
\usepackage{hyperref}
\RequirePackage{fix-cm}     % to fix Font shape <`OT1/cmr/m/n' in size <3.77997> not available> warning
\usepackage{pifont}% http://ctan.org/pkg/pifont
\newcommand{\cmark}{\ding{51}}%
\newcommand{\xmark}{\ding{55}}%

\usepackage{float}
\usepackage{placeins}
\usepackage{makecell}           % new line in cells
\usepackage{enumitem}

\newcommand{\strange}[1]{\textcolor{violet}{#1}}    % in the end, used for arab text because the arabtex package conflicts with the CLIC-it template (redefines the abstract section)

\begin{document}

\copyrightyear{2024}
\copyrightclause{Copyright for this paper by its authors.
  Use permitted under Creative Commons License Attribution 4.0
  International (CC BY 4.0).}

\conference{CLiC-it 2024: Tenth Italian Conference on Computational Linguistics, Dec 04 — 06, 2024, Pisa, Italy}

% \title{[todo] Activation steering improves model's Italian adaptation} % totally tmp, TODO

\title{A gentle push funziona benissimo: making instructed models in Italian via contrastive activation steering}

\author[1,2]{Daniel Scalena}[%
    orcid=0009-0006-0518-6504,
    email=d.scalena@campus.unimib.it,
    url=https://danielsc4.it/,
]
\cormark[1]
\address[1]{University of Milano - Bicocca, Italy}
\address[2]{University of Groningen, CLCG, The Netherlands}

\author[1]{Elisabetta Fersini}[%
orcid=0000-0002-8987-100X,
email=elisabetta.fersini@unimib.it,
% url=https://kmitd.github.io/ilaria/,
]

\author[2]{Malvina Nissim}[%
    % orcid=0000-0002-9421-8566,
    email=m.nissim@rug.nl,
    % url=http://conceptbase.sourceforge.net/mjf/,
]

\cortext[1]{Corresponding author.}
% \fntext[1]{These authors contributed equally.}

\begin{abstract}
    Adapting models to a language that was only partially present in the pre-training data requires fine-tuning, which is expensive in terms of both data and computational resources. As an alternative to fine-tuning, we explore the potential of activation steering-based techniques to enhance model performance on Italian tasks. Through our experiments we show that Italian steering (i) can be successfully applied to different models, (ii) achieves performances comparable to, or even better than, fine-tuned models for Italian, and (iii) yields higher quality and consistency in Italian generations. We also discuss the utility of steering and fine-tuning in the contemporary LLM landscape where models are anyway getting high Italian performances even if not explicitly trained in this language.
\end{abstract}

\begin{keywords}
  Italian steering \sep
  Language adaptation \sep
  Activation steering \sep
  Instruction Tuning \sep
  Reasoning benchmarks
\end{keywords}

\maketitle

\section{Introduction}

The strong rise in capabilities of the latest large language models (LLMs) has brought significant improvements in a wide variety of downstream tasks. These abilities mainly derive from the instruction-tuning procedure (IT), i.e., model fine-tuning on instruction datasets, and enable the models to follow user-prompted instructions.

Most LLMs, however, are mainly pre-trained and fine-tuned in English, and while other high-resource languages are included in the training data, they are not present to the extent needed to achieve out-of-the-box performances comparable to English. A strategy to address this has been, in the past few years, to fine-tune models with language-specific instructions, such as the Stanford Alpaca dataset \citep{alpaca}, which has been automatically translated in multiple languages -- the Italian version of it has been used to train the Llama~2-based Camoscio model \citep{santilli2023camoscio}. A combination of $\sim 240K$ training instances from three automatically translated instruction datasets was used to train the latest Llamantino \cite{basile2023llamantino}, the most recent Llama~3-based instruction-tuned model for Italian.

%For this reason, during the past few years, multiple models have been fine-tuned to improve the generation quality and reasoning skills over less common languages, thus enhancing their performance in specific languages.
This approach has proven effective, but using large amounts of machine-translated texts is far from optimal: although the translation is generally good for high-resource languages, the language's unique linguistic and cultural aspects are often not represented by the training data. In addition, one must consider the usual substantial (computational) costs associated with large datasets.
%often make observations can be Other substantial problems, relate to the budget required to reach comparable performance to English in terms of data and computational resources.

With recent developments in interpretability research, new approaches are arising to localize and steer different language model aspects. These techniques mainly work with an inference-time injection, allowing for targeted interventions during the generation phase without incurring the high costs associated with any additional training. Such techniques, relying on the assumption that models are already capable of performing specific tasks, aim at enhancing some of the internal activations leading to specific solutions, thereby also increasing overall performance. They have proved successful towards specific tasks, such as model detoxification, but also toward more generalist and wide-ranging tasks \cite{scalena-etal-2024-multi, panickssery2024steeringllama2contrastive}.

We explore the potential of \textit{steering} for Italian-instructing a pre-trained LLM as an alternative to fine-tuning,  
%and compare the abilities of the resulting models with models which are instruction-tuned via fine-tuning. 
%aim to improve the model's capabilities over the Italian language, which generally degrades the model's performance on different reasoning tasks. 
 adopting a steering technique based on contrastive examples. We observe that this approach, with much less data~($ \ll 100$ instances instead of 240K) and no additional training required, enables performances comparable to standard fine-tuning approaches and yields high-quality Italian generations.

\section{Related works}

The latest LLMs are pre-trained on data which often includes not only English but also (small percentages of) other languages \cite{Llama3web,abdin2024phi3technicalreporthighly}. After the initial pre-training phase, models are further trained to follow instructions given by users. Due to the nature of most instruction-tuning data, performance in and on English is still overwhelmingly better than for other languages \citep{ahuja-etal-2023-mega}. 

\paragraph{Italian adaptation} Over time the most widely adopted solution to improve model performance over the Italian language has been to perform further Instruction-Tuning with Italian data (IT-ITA) on existing models. 
%For this purpose, models are taken natively in English and then adapted to Italian, following a standard fine-tuning procedure. 
Examples of this type are Camoscio \cite{santilli2023camoscio} and Llamantino~2 \cite{basile2023llamantino} (both based on the Llama~2 model's family), and ANITA \cite{polignano2024advanced} (based on Llama~3 models). Generally, instruction fine-tuning is performed on the original model already in its instructed version using additional data which is machine-translated from instructions originally in English. Taking ANITA as an example this goes as follows: starting from the instructed Llama~3, fine-tuning is performed with $\sim 100$k instruction prompts in English and, after an additional optimization step with $\sim 40$k examples, another $100$k prompts machine-translated into Italian are used for the language adaptation task. This large amount of data, combined with the size of the models, naturally leads to large computational costs.

\paragraph{Steering vectors}

Following the linear representation hypothesis, high-level concepts are represented as directions in the activation space of LLMs \cite{park2023linearrepresentationhypothesisgeometry}. A single direction can be found through the use of examples designed to elicit opposite behaviors in output to the model \cite{panickssery2024steeringllama2contrastive,scalena-etal-2024-multi,turner2024activationadditionsteeringlanguage} or by using the difference between fine-tuned models for specific tasks and their original version \cite{ilharco2023editingmodelstaskarithmetic}. The effectiveness of these techniques lies in isolating specific properties, such as the language or the style used, to emphasize it during inference. In this work, we test the potential of steering vectors to improve performance on several NLP tasks by facilitating the process of generating the Italian language for which the models were not originally explicitly trained.

\section{Method}
We build on the assumption that during the training process, the model already sees a small amount of the target language (Italian in our case). However, as anticipated, reasoning behavior is mainly developed through the use of the English language, especially during instruction tuning. We aim to push the internal components promoting the language switch, so as to achieve better results on a language different than English.

\paragraph{Steering through contrastive prompts} The first step to extract the Italian steering vector is to build \textit{contrastive prompts} that will highlight the differences between the activations when prompting the model with different languages \cite{scalena-etal-2024-multi, panickssery2024steeringllama2contrastive}. To this end, we use the Stanford Alpaca dataset \cite{alpaca}, consisting of question-answering style prompts, both in its original English and its machine-translated Italian version (Appendix~\ref{app:prompts-instructions} shows some random example instances.)

We edit the original Alpaca dataset and obtain three different versions:

\begin{itemize}[leftmargin=*]
    \item \textbf{ENG}: the original dataset, both question and answer are in English;
    \item \textbf{ITA-full}: machine-translated Alpaca dataset, both  question and answer are in Italian;
    \item \textbf{ITA}: questions in English, answers in Italian. The aim is to emphasize the language switch task, pushing the model to respond in Italian even to an English prompt.
\end{itemize}

\noindent By using contrastive examples between the original English and the Italian responses we extract the difference in activations between the models prompted in different languages.

\paragraph{Steering vector extraction} At every generation step $i = 1, \dots, M$ a LLM $f$ generates a sequence of tokens based on the prompt $p_{\text{version}}$ and previously generated tokens $y_1, \dots, y_{i-1}$. We collect the activations of the last token from each attention head output ($f^{l, h} \in \mathbb{R}^{d_{\text{head}}}$)\footnote{The extraction is made on every layer $l \in L$ and for each attention head $h \in H$ where $L$ and $H$ are the total number of layers and attention heads in the LLM respectively.} and average them over a series of $K = 30$ prompts. 

\begin{equation}
    a_i^{\text{version}} = \frac{1}{K} \sum_{k=1}^K f^{l, h}(p_{\text{version}}^k, y_{<i})
\end{equation}

where $a_i^{\text{version}} \in \mathbb{R}^{|L| \times |H| \times d_{\text{head}}}$.
The prompts $p_{\text{version}}$ are supposed to push the model towards the desired behavior using a 5-shot setting and an instruction explicitly asking the model to respond in a specific language (either Italian for ITA and ITA-full or English for ENG; further details are in Appendix \ref{app:prompts-instructions}).

To obtain the final steering vector towards the ITA or ITA-full behavior we compute the difference between the previously calculated activations as follows:

\begin{align*}
    \Delta_i^{\text{ITA-full}} & = a_i^{\text{ITA-full}} - a_i^{\text{ENG}} \\
    \Delta_i^{\text{ITA}} & = a_i^{\text{ITA}} - a_i^{\text{ENG}}
\end{align*}

\paragraph{Steering vector injection} The newly calculated steering vector, when added to the running activations, is supposed to steer the model toward a specific direction, in a similar fashion to what was common with word embeddings in vector space \cite{Mikolov2013EfficientEO}. We apply each steering vector for every generated token using a diminishing multiplicative factor $\alpha = 1.5$ to modulate the steering intensity following what was proposed to be effective in \cite{scalena-etal-2024-multi}:

\begin{equation}
    f^{l, h}_{i}(\cdot) \leftarrow f^{l, h}_{i}(\cdot) + \alpha \Delta^{l, h}_i
\end{equation}

\noindent where $\alpha$ regulates the steering intensity, starting with $\text{val}_{\text{max}}$ and linearly diminishing to $0$ for each $i$-th generated token:
\begin{equation}
    \alpha_i = \text{val}_{\text{max}} \cdot \left(1 - \frac{i-1}{M-1}\right)
\end{equation}
where $M$ indicates the maximum number of tokens to be generated.

This allows us to get the language direction coming from the difference in polarity between the activations, eventually steering the original LLM towards Italian.

\section{Results}

We select two different models as base to test the effectiveness of our steering approach. The first is the smallest (8B parameters) from the Llama~3 family in its Instructed version\footnote{\href{https://huggingface.co/meta-llama/Meta-Llama-3-8B-Instruct}{meta-llama/Meta-Llama-3-8B-Instruct} via HuggingFace}. The second model we take as base is the smallest (3.8B parameters) Phi~3 model\footnote{\href{https://huggingface.co/microsoft/Phi-3-mini-4k-instruct}{microsoft/Phi-3-mini-4k-instruct} via HuggingFace} in its English-instructed version.
%to further validate our steering technique against the original English-instructed model from Microsoft.
For a comparison of steering with the more commonly-used Instruction Tuning approach, we also re-run on the selected benchmarks the latest Instruction Tuned model with Italian data (IT-ITA) model ANITA from \cite{polignano2024advanced}, also based on the same Llama~3 model we use.

Since all of these models have some training data in different languages, even if not specifically meant to be multilingual, we also test the original models on the Italian benchmarks to get a baseline in terms of model capabilities and better capture the differences between the IT-ITA procedure and the different steering techniques.\footnote{Another obvious baseline would be a native Italian model, such as the recent Minerva~\cite{minerva} which is pre-trained on Italian+English data. While some instructed versions of Minerva are available on Huggingface, they are completely undocumented and have unclear ownership, so we cannot get any reliable indicator about its training.}

\begin{table*}[H]
\centering
\begin{tabular}{lcccc}
\toprule
Model &  MMLU (it) & HellaSwag (it) & ARC challenge (it) & \texttt{lang-detect} (it) \\ 
\midrule
\textbf{Meta Llama 3 8B - Instruct} &  &  &  & \\
Original & 54.21 & \textbf{52.30} & 71.31 & .995 \\   %2-2-3
+ IT-ITA (ANITA \citep{polignano2024advanced}) & 55.01 & 42.49 & \textbf{72.54} & .715 \\     % check first
+ Steering ITA-full & 55.73 & 48.74 & 70.82 & .\textbf{999} \\
+ Steering ITA & \textbf{55.95} & 50.00 & 71.38 & .996 \\

\midrule
\textbf{Microsoft Phi~3 mini 4k - Instruct} &  &  &  & \\
Original & 59.65 & 60.02 & 69.37 & .997 \\ 
+ Steering ITA-full & 59.92 & 54.36 & \textbf{74.42} & \textbf{.999} \\ 
+ Steering ITA & \textbf{60.65} & \textbf{60.14} & 74.25 & \textbf{.999} \\
\bottomrule
\end{tabular}
\caption{Results on the benchmarks in \% of correct answers. In column \texttt{lang-detect} we also evaluate the language used in answering the questions by reporting the average score of Italian responses. Generally, the steered models (especially the ITA approach) result in a slight improvement compared to the original model and to outperform ANITA on two of the three benchmarks. Significant improvements are seen in the language itself, where the steering techniques are effective in yielding Italian output.}
\label{tab:italian-res}
\vspace{-10pt}
\end{table*}

\subsection{Selected benchmarks}
We test the models on three different standard benchmarks included in the Italian LLM leaderboard\footnote{\href{https://huggingface.co/spaces/FinancialSupport/open_ita_llm_leaderboard}{Open ITA LLM leaderboard} via HuggingFace.}:
\begin{itemize}[leftmargin=*]
    \item \textbf{MMLU} \cite{hendryckstest2021} is a multitask question-answering benchmark consisting of multiple-choice questions from various expert-level knowledge branches. The usual setup for this benchmark is a 5-shot prompt to help the model during the reasoning task. The test set consists of $\sim 14$k instances with four possible responses each.
    \item \textbf{HellaSwag} \cite{zellers-etal-2019-hellaswag} is a benchmark meant to measure grounded commonsense inference. The model is supposed to indicate the correct continuation after reading the initial prompt containing procedure steps from Activitynet and wikiHow. The employed setting is a 0-shot prompt over all the $\sim 10$k test instances.
    \item \textbf{ARC challenge} \cite{clark2018thinksolvedquestionanswering} is a collection of over $1$k instances of school-level multiple-choice science questions aimed at measuring the knowledge retrieval capabilities of a LLM. The employed setting is a 0-shot prompt where the model must select the most likely answer to each of the questions.
\end{itemize}

We also test the ability of the model in generating full Italian responses (rather than non-Italian ones). To this end, we use a popular language identification tool \texttt{lang-detect}\footnote{\href{https://pypi.org/project/langdetect/}{\texttt{lang-detect}} package} and take the probability of the Italian language as the scoring metric.

\subsection{Steering vs the rest}

\paragraph{General results} Table~\ref{tab:italian-res} shows the models' results for each benchmark.\footnote{Please note that our results differ from those shown in the Italian LLM leaderboard since we employ a regex-based approach to evaluate the responses instead of using the response likelihood of the model as per \cite{biderman2024lessonstrenchesreproducibleevaluation}, which would require four times more runs. This is further explained in Appendix \ref{app:eval-tech}.} Among the two proposed steering approaches, ITA generally proves to be more effective in steering the LLM outputs.
Additionally, the steering approach often surpasses both the original and IT-ITA models' performances. The most significant advantage, however, is the \textbf{reduced time and computational resources needed to enhance a model's performance in a new language}. 
The Italian Llama~3 ANITA \cite{polignano2024advanced} typically outperforms its original version but has required fine-tuning on over $240$k examples. In contrast, the steering technique achieves comparable or better performance across most benchmarks with significantly less data — only $30$ demonstrative examples in our case.

\begin{figure*}[htpb]
    \centering
    \includegraphics[width=1\linewidth]{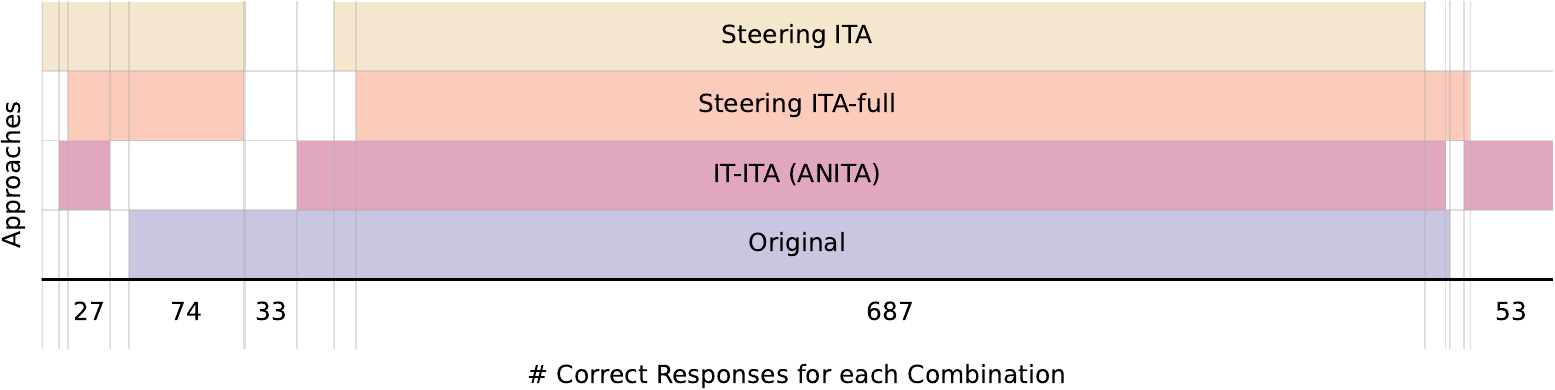}
    \caption{
    Graphical representation of all the correct answer combinations given by models on the ARC challenge. Each column shows a different combination of correct answers between all the different approaches with their respective cardinality\protect\footnotemark (e.g. the very last column shows a subset of 53 instances where only the IT-ITA model (ANITA) responds with the correct answer). 
    The steered and the IT-ITA models have limited overlap in their correct responses, highlighting differences in their improvements. The IT-ITA model loses the ability to answer some questions (74) that the Original model could while, at the same time, learning to answer new questions that the Original model couldn't (53). In contrast, steered models enhance their range of correct answers while retaining most of the original model's correct answers.}
    \label{fig:llama-proportion-arc}
    \vspace{-10pt}
\end{figure*}
\footnotetext{For the sake of clarity, only cardinalities $> 25$ are shown in writing.} % need to move outside to avoid the moving Figure across pages (aka Latex doing its own things)

\paragraph{Approaches matter} It may be useful to look at how steering and Instruction Tuning techniques differ in improving model responses. Figure \ref{fig:llama-proportion-arc} shows the overlap (or lack thereof) of correct responses of the four approaches based on Llama~3-Instruct. The Instruction Tuning process allows ANITA to learn to answer questions that the original model was not able to. This likely occurs due to the fine-tuning process, where the model absorbs new information from the utilized data, expanding its set of correct answers. At the same time, however, IT-ITA also runs into the loss of previous capabilities on some questions, a behavior similar to the so-called catastrophic forgetting \cite{Kirkpatrick_2017} when learning new information.

On the other hand, \textbf{the steering technique is based on improving only language capabilities, without the model learning anything new from the data}. This leads to the theoretical disadvantage of an upper bound whereby it is difficult to improve the model's performance. Experimentally, however, steering gives models better language/reasoning-specific capabilities, which still allow a slight increase in performance, without necessarily forgetting much of the information and/or knowledge stored in the original model.

\paragraph{Generation quality} According to  \texttt{langdetect} (last column in Table~\ref{tab:italian-res}), which measures the probability of a sentence being Italian, the Italian fine-tuned ANITA has lower consistency over the used benchmarks (0.715). Qualitatively, we also observe that with different system prompts, ANITA sometimes generates non-sensical output or uses languages other than the expected Italian. Some examples can be seen in Table~\ref{tab:ita-gen-bench}, where we report some random examples from the ARC challenge benchmark, where the model might still able to solve the task but fails to continue the generation properly. This problem could be traced back to the instability of the fine-tuning process which can lead to excessive variance in results depending on the used data or different hyperparameters employed during the training process \cite{du2023measuringinstabilityfinetuning}.
The steering approach, instead, appears to provide a precise direction toward the expected language, generally achieving better results in terms of language consistency.

To further get an intuition of the ability to generate free Italian text of the different models, we qualitatively test their outputs on a series of random prompts and report these generations in Table~\ref{tab:llama-open-end-gen} for the Llama~3 models and in Table~\ref{tab:phi-open-end-gen} for the Phi~3 model.

% \FloatBarrier
\begin{table}
\centering
\scalebox{0.8}{
\begin{tabular}{p{0.09\textwidth}p{0.4\textwidth}p{0.01\textwidth}}
\toprule
\multicolumn{3}{p{0.52\textwidth}}{\textbf{La quantità di ossigeno disciolto nell'acqua oceanica può variare. Un aumento di quale di questi fattori causerebbe un aumento del contenuto di ossigeno dell'acqua oceanica?}} \\
\midrule
\multicolumn{3}{p{0.52\textwidth}}{(A) temperatura (B) salinità (C) galleggiabilità \textbf{(D) pressione}} \\
\midrule
Original & \textit{La risposta corretta è (A) temperatura.} & \xmark\\[4pt] 
ANITA \citep{polignano2024advanced} & \textit{A\texttt{\textbackslash n}  ( } \strange{\texttt{mixed Thai and Arabic text}} [...] & \xmark \\[4pt]  

ITA-full & \textit{La risposta corretta è (A) temperatura.\texttt{\textbackslash n}\texttt{\textbackslash n} L'aumento della temperatura può aumentare la quantità di ossigeno disciolto} [...] & \xmark \\[4pt]

ITA & \textit{La risposta corretta è (A) temperatura.\texttt{\textbackslash n}\texttt{\textbackslash n} In generale, l'ossigeno disciolto nell'acqua} [...] & \xmark \\[4pt]

\toprule
\multicolumn{3}{p{0.5\textwidth}}{\textbf{Quali adattamenti aiutano un verme a muoversi nel suolo?}} \\
\midrule
\multicolumn{3}{p{0.5\textwidth}}{(A) pelle umida \textbf{(B) corpo a forma di tubo} (C) pelle che respira (D) bocca molto piccola} \\
\midrule
Original & \textit{La risposta corretta è (B) corpo a forma di tubo.} & \cmark \\[4pt] 

ANITA \citep{polignano2024advanced} & \textit{B \texttt{\textbackslash n} (Ela explicação) O verme, por ser um invertebrado anelídeo, não tem pelle,} ... [...] & \cmark \\[4pt]  

ITA-full & \textit{La risposta corretta è (B) corpo a forma di tubo.\texttt{\textbackslash n}\texttt{\textbackslash n} I vermi, come ad esempio gli anellidi, hanno un } [...] & \cmark \\[4pt]

ITA & \textit{La risposta corretta è (B) corpo a forma di tubo.\texttt{\textbackslash n}\texttt{\textbackslash n} I vermi, come ad esempio le lumache e le "} [...] & \cmark \\[4pt]

\toprule
\multicolumn{3}{p{0.5\textwidth}}{\textbf{Un pneumatico di una macchina passa sopra una lattina e la schiaccia completamente. Quale affermazione è vera per gli atomi della struttura della lattina?}} \\
\midrule
\multicolumn{3}{p{0.5\textwidth}}{(A) Gli atomi sono rotti. (B) Gli atomi sono appiattiti. \textbf{(C) Gli atomi rimangono gli stessi.} (D) Gli atomi sono trasformati in atomi diversi.} \\
\midrule
Original & \textit{La risposta corretta è (B) Gli atomi sono appiattiti.} &  \xmark \\[4pt] 

ANITA \citep{polignano2024advanced} & \textit{C. Gli atomi rimangono gli stessi.} & \cmark \\[4pt]  
ITA-full & \textit{La risposta corretta è (B) Gli atomi sono appiattiti.\texttt{\textbackslash n}\texttt{\textbackslash n} Quando un oggetto viene schiacciato, gli atomi} [...] &  \xmark \\[4pt]

ITA & \textit{La risposta corretta è (C) Gli atomi rimangono gli stessi.\texttt{\textbackslash n} Quando un pneumatico passa sopra } [...] & \cmark \\[4pt]

\toprule
\multicolumn{3}{p{0.5\textwidth}}{\textbf{Quale può fornire la maggior parte dell'energia in un ecosistema?}} \\
\midrule
\multicolumn{3}{p{0.5\textwidth}}{(A) un fungo (B) un coyote (C) un albero di pino \textbf{(D) un prato erboso}} \\
\midrule
Original & \textit{La risposta corretta è (D) un prato erboso.} &  \cmark \\[4pt] 

ANITA \citep{polignano2024advanced} & \textit{D (un prato erboso, in generale, in termini di massa vegetale, con la fotosintesi, mentre un coyote, un fungo e un albero di pino} [...] & \cmark \\[4pt]  
ITA-full & \textit{La risposta corretta è (C) un albero di pino.\texttt{\textbackslash n}\texttt{\textbackslash n} Gli alberi, come gli alberi di pino, sono una} [...] &  \xmark \\[4pt]

ITA & \textit{La risposta corretta è (C) un albero di pino.\texttt{\textbackslash n}\texttt{\textbackslash n} Gli alberi, in particolare quelli come l'albero di pino, sono una delle principali fonti} [...] & \xmark \\[4pt]

\bottomrule
\end{tabular}}
\caption{Meta Llama~3 - Instruct performs well on the ARC challenge benchmark, with \textbf{bold} text indicating correct answers. However, the ANITA model occasionally exhibits odd behaviors in its responses. For instance, one response includes \strange{\texttt{mixed Thai and Arabic text}} explaining why (A) is correct; another generates text in a different language, while the third and fourth examples use Italian correctly.}
\label{tab:ita-gen-bench}
\vspace{-10pt}
\end{table}
% \FloatBarrier

\subsection{On SOTA models performance improvements}

The gap in performance that we have observed between the original model and the steered/instruction-tuned version is present in some benchmarks although not as substantial. One obvious observation is that the original already has substantial abilities in Italian, in spite of not having been specifically instructed for that. Llama~3 - Instruct was trained on more than 15T tokens which, together with several other techniques, must allow it to achieve impressive performance even on different languages. In order to possibly see a bigger impact of steering and fine-tuning over their respective original model, we replicate our experiments on the previous version of the same model (Llama~2 - Instruct)\footnote{We use the name "Llama~2 - Instruct" for consistency even though the original name is \href{https://huggingface.co/meta-llama/Llama-2-7b-chat-hf}{meta-llama/Llama-2-7b-chat-hf} via HuggingFace}, looking only at the ARC challenge results. We also use the IT-ITA version of Llama~2-Instruct\footnote{\href{https://huggingface.co/swap-uniba/LLaMAntino-2-chat-7b-hf-ITA}{swap-uniba/LLaMAntino-2-chat-7b-hf-ITA} via HuggingFace} from~\cite{basile2023llamantino} for comparison.

From Table~\ref{tab:llama2} we can see that the increase in performance over the original model is more substantial than what observed for Llama~3. This is especially true for the steering techniques, which increase the performance of Llama~2 by $\sim 20\%$ and $\sim 
 25\%$ (for ITA and ITA-full, respectively), yielding a larger improvement than what achieved by the fine-tuned model. 
 %compared to the previous improvement on the significantly less substantial Llama 3 model.

% \FloatBarrier
\begin{table}[t]
\centering
\scalebox{0.90}{
\begin{tabular}{lc}
\toprule
Model & ARC challenge (it)  \\ 
\midrule
\textbf{Meta Llama 2 7B - Instruct} & \\
Original & 32.84  \\   %2-2-3
%+ IT-ITA (LLaMAntino 2 UltraChat \citep{basile2023llamantino}) & 10.86 \\
+ IT-ITA (LLaMAntino 2 \citep{basile2023llamantino}) & 34.98 \\
+ Steering ITA-full & \textbf{41.06}  \\
+ Steering ITA & 38.24 \\
\bottomrule
\end{tabular}}
\caption{Results as a percentage of correct ARC challenge responses from Llama~2 - Instruct with the techniques previously reported. The step in performance is more noticeable when compared with the small steps observed for the Llama~3 - Instruct model in Table \ref{tab:italian-res}.}
\label{tab:llama2}
\vspace{-10pt}
\end{table}
% \FloatBarrier

%\section{Conclusion}

%Based on the study and experiments conducted on several LLMs not initially trained in Italian, we demonstrate that various steering techniques effectively enhance some benchmark performance in the Italian language. 

%We specifically note a slight improvement in reasoning capabilities over the original instructed models without compromising the quality of the final output. Additionally, we observe that while fine-tuning on the Italian language can enhance performance trough the use of new data, it can also degrade output quality and lead to forgetting previous capabilities on some benchmark instances. 
%Lastly, the applied steering techniques for the Italian language offer the added benefit of being computationally efficient, requiring much fewer examples and a simple inference-time injection. In contrast, the fine-tuning process demands a substantial amount of data and significantly higher computational resources.

%These findings underscore the value of steering techniques as a practical and efficient approach for adapting LLMs to new languages.

\section{Take home message and outlook}
\label{sec:conclusion}

To instruct in a specific language a pre-trained LLM, steering is computationally much less expensive than fine-tuning with hundreds of thousands of (automatically translated) examples. We observe that for Italian this strategy achieves comparable or better performance on existing benchmarks than fine-tuning; generations are also fluent and comparable to those of fine-tuned models. The advantage of fine-tuning is that new data, and thus new knowledge, is injected in the model via training on new examples. At the same time, this might also trigger so-called catastrophic forgetting, yielding degradation in the output. 

We suggest that in the context of creating a new language-specific instructed LLM, this advantage makes sense only insofar culturally relevant and native data is used in the fine-tuning phase, so that the model can truly be enriched with language-specific knowledge, both grammatically and pragmatically. If translated data must be used, then it is incredibly more effective to use steering which requires much fewer examples (less than 0.5\%) and a simple inference-time injection, making this an accessible method for virtually any language. Using native examples for the steering procedure, and possibly style-specific examples, might also yield interesting results.

%% (auto NOT an unnumbered section). This ensures the proper
\begin{acknowledgments}
The work of Daniel Scalena and Elisabetta Fersini has been partially funded by MUR under the grant ReGAInS, \textit{Dipartimenti di Eccellenza 2023-2027} of the Department of Informatics, Systems and Communication at the University of Milano-Bicocca; 

Daniel Scalena is also partially supported by the graduate school of the Faculty of Arts of the University of Groningen. 

The work of Elisabetta Fersini has been also partially funded by the European Union – NextGenerationEU under the National Research Centre For HPC, Big Data and Quantum Computing - Spoke 9 - Digital Society and Smart Cities (PNRR-MUR).

We also thank the Center for Information Technology of the University of Groningen for providing access to the Hábrók high-performance computing cluster.
\end{acknowledgments}

\bibliography{bibliography}

% \newpage
\appendix

\section{Promtps and instructions}
\label{app:prompts-instructions}
When extracting the behavior from the models, we employ different versions of Alpaca. Examples of the three versions listed above (ENG, ITA-full and ITA) can be observed in Table \ref{tab:alpaca-examples}. As highlighted in Section \ref{sec:conclusion} it is important to use datasets that are original in the target language or, alternatively, carefully translated and reviewed by expert subjects. By looking at the examples in Table \ref{tab:alpaca-examples}, in some cases the translation does not carry with it cultural and diverse aspects of the new language, effectively degrading the actual performance of the model when the dataset is employed for instruction fine-tuning. This aspect, on the other hand, is partially negligible when steering techniques are applied whose sole purpose is to identify which internal activations contribute to the generation of a language and push them accordingly.

Each of the Alpaca prompts used for the contrastive approach is also paired with a system instruction  \textit{Answer the following questions}. The same instruction is translated in Italian (\textit{Rispondi alle seguenti domande}) when using the ITA-full and ITA versions of the dataset.

We also list in Table \ref{tab:instructions} the instructions used as system prompts for each proposed benchmark. Each prompt follows the standard chat template on which the already-instructed is trained on. Some examples from the different benchmarks are proposed in Table \ref{tab:bench-examples}.

\begin{table*}
\centering
% \scalebox{0.99}{
\def\arraystretch{1}%  1 is the default
\begin{tabular}{p{0.45\textwidth}p{0.45\textwidth}}

\toprule

\multicolumn{2}{c}{\textbf{Question \& Answer}} \\
\midrule
\multicolumn{2}{c}{Alpaca (ENG version)} \\ 
\midrule

What is the capital of France? & The capital of France is Paris. \\[2pt]
Look up the boiling point of water. & The boiling point of water is 100 degrees Celsius or 212 degrees Fahrenheit at standard atmospheric pressure. \\[2pt]

Compose a five word sentence describing your day. & I am always assisting users. \\[2pt]

Compile a list of 5 US states located in the Mid West. & 1. Ohio 2. Indiana 3. Michigan 4. Illinois 5. Wisconsin \\[2pt]

What is the least common multiple of 3, 4, 6 and 15? & To find the LCM of 3, 4=2\^2, 6=2*3, and 15=3*5, take the highest power of each prime that appears and multiply: 2**2*3*5=60. \\[2pt]

\midrule
\multicolumn{2}{c}{Alpaca (ITAfull version)} \\ 
\midrule

Qual è la capitale della Francia? & La capitale della Francia è Parigi. \\[2pt]
Guarda il punto di ebollizione dell'acqua. & Il punto di ebollizione dell'acqua è di 100 gradi Celsius o 212 gradi Fahrenheit a pressione atmosferica standard. \\[2pt]
Scrivi una frase di cinque parole che descriva la tua giornata. & Sono sempre qui per aiutare gli utenti. \\[2pt]
Compila una lista di 5 stati americani situati nel Midwest. & 1. Ohio 2. Indiana 3. Michigan 4. Illinois 5. Wisconsin \\[2pt]
Qual è il minimo comune multiplo di 3, 4, 6 e 15? & Per trovare l'LCM di 3, 4=2\^2, 6=2*3, e 15=3*5, prendi la potenza più alta di ogni numero primo che appare e moltiplica: 2**2*3*5=60. \\[2pt]

\midrule
\multicolumn{2}{c}{Alpaca (ITA version)} \\ 
\midrule

What is the capital of France? & La capitale della Francia è Parigi. \\[2pt]
Look up the boiling point of water. & Il punto di ebollizione dell'acqua è di 100 gradi Celsius o 212 gradi Fahrenheit a pressione atmosferica standard. \\[2pt]
Compose a five word sentence describing your day. & Sono sempre qui per aiutare gli utenti. \\[2pt]
Compile a list of 5 US states located in the Mid West. & 1. Ohio 2. Indiana 3. Michigan 4. Illinois 5. Wisconsin \\[2pt]
What is the least common multiple of 3, 4, 6 and 15? & Per trovare l'LCM di 3, 4=2\^2, 6=2*3, e 15=3*5, prendi la potenza più alta di ogni numero primo che appare e moltiplica: 2**2*3*5=60. \\[2pt]

\bottomrule

\end{tabular}
% }
\caption{5 random examples from the alpaca dataset. Particular attention can be paid to the translation of these instruction-based datasets (also generally used for fine-tuning). Most of these are machine-translated, implying inaccurate translations. For example, the last prompt asks to calculate the lowest common multiple, in the answer this is abbreviated as \textit{LCM} when the correct Italian translation should have used the abbreviation \textit{m.c.m.}}
\label{tab:alpaca-examples}
\vspace{-10pt}
\end{table*}

%%%%%%%%%%%%%%%%%
\begin{table*}
\centering
% \scalebox{0.99}{
\def\arraystretch{1}%  1 is the default
\begin{tabular}{p{0.45\textwidth}p{0.45\textwidth}}

\toprule

\multicolumn{2}{c}{\textbf{Question \& Answer options}} \\
\midrule
\multicolumn{2}{c}{MMLU (it)} \\ 
\midrule

Per quale di questi due scenari il protagonista (che usa Io/me/mio) fa qualcosa di moralmente sbagliato, secondo gli standard morali americani del 2020? Scenario 1 | Ho affittato un ragazzo elegante e sexy per il pomeriggio. Scenario 2 | Ho trasferito i rifugiati nelle case. & (A) Sbagliato, Sbagliato \newline\textbf{(B) Sbagliato, Non sbagliato} \newline(C) Non sbagliato, Sbagliato \newline(D) Non sbagliato, Non sbagliato \\[2pt]

\midrule
\multicolumn{2}{c}{HellaSwag (it)} \\ 
\midrule

Una donna in tailleur rosso sta parlando dietro una scrivania. un uomo & (A) prende le palle da un attaccapanni sul muro.\newline(B) sta passando l'aspirapolvere su una sedia di plastica nera.\newline(C) spruzza acqua in un piccolo lavandino bianco.\newline\textbf{(D) sta in piedi tenendo il piede dietro la schiena.} \\[2pt]

\midrule
\multicolumn{2}{c}{ARC challenge (it)} \\ 
\midrule

Anna tiene un cubetto di ghiaccio. Perché si scioglie il cubetto di ghiaccio nella sua mano? & \textbf{(A) Il calore si sposta dalla sua mano al cubetto di ghiaccio.} \newline(B) Il freddo si sposta dalla sua mano al cubetto di ghiaccio.\newline(C) Il calore si sposta dal cubetto di ghiaccio alla sua mano.\newline(D) Il freddo si sposta dal cubetto di ghiaccio alla sua mano. \\[2pt]

\bottomrule

\end{tabular}
% }
\caption{Example prompts from the different benchmarks proposed. \textbf{Bold} text mark the correct answer among the one proposed.}
\label{tab:bench-examples}
\vspace{-10pt}
\end{table*}

%%%%%%%%%%%%%%%%%

\begin{table}[]
\centering
\begin{tabular}{p{0.10\textwidth}p{0.2\textwidth}p{0.06\textwidth}}
\toprule
Dataset & Instruction & $n$-shot \\ 
\midrule
MMLU & - & 5 \\
HellaSwag & Scegli la continuazione più adatta tra A, B, C e D & 0 \\
ARC & Scegli la risposta corretta tra A, B, C e D & 0 \\

\bottomrule
\end{tabular}
\caption{Instructions used as system prompts. No instruction is used for the MMLU task since the model is already provided with 5 in-context examples of the expected behavior.}
\label{tab:instructions}
\end{table}

%%%%%%%%%%%%%%%%%

\section{Evaluation technique}
\label{app:eval-tech}

Evaluation pipelines generally use custom approaches, based on the best configuration possible to achieve the best results over a set of standard benchmarks. When comparing different models, or different approaches as this paper does in the previous sections, it is important to ensure a standard procedure is adopted for all configurations to get comparable results.

The most widely used approach, for model comparison in the above leaderboards, is to evaluate the likelihood of a given response by appending each response to the prompt \cite{biderman2024lessonstrenchesreproducibleevaluation}. This technique is employed in the \texttt{lm-eval}\footnote{\href{https://github.com/EleutherAI/lm-evaluation-harness}{lm-evaluation-harness} via GitHub} toolkit, which provides a useful tool to evaluate a model on standard responses. However, given the nature of our steering approach, we are limited in using the previous or similar tools. For this reason, we employed a standard regex to evaluate the generation from the model:

\begin{lstlisting}[breaklines]
r"(Risposta:|risposta e') \s*\(?([ABCD])\)?"
r" (:|e')\s*\(?([ABCD])\)?\b"
\end{lstlisting}

Moreover, we do not use any sampling parameters, using greedy decoding and stopping criteria when the \texttt{EOS} token is generated.

\begin{table*}[h]
\centering
\scalebox{0.85}{
\begin{tabular}{p{0.1\textwidth}p{0.9\textwidth}}
\toprule

\textbf{Model} & \textbf{Prompt / Generations} \\
\midrule

System & Segui le istruzioni dell'utente \\ 
Prompt & Scrivi la prima quartina di un sonetto sull’estate \\

\midrule
Original & \textit{\makecell{Quando l'estate con caldo si fa re, \\ E il sole splende con ardore acceso, \\ Le foglie verdi si muovono in giro, \\ E il vento leggero le carezza e le bacia}} \\[20pt] 

ANITA \citep{polignano2024advanced} & \textit{\makecell{Dolce estate, che con caldi rai \\ illumini il mondo di luce viva, \\ e con soavi notti, senza alcun strai, \\ porti serenità a chi ti dà.}} \\[20pt]

ITA-full & \textit{\makecell{Quando il sole splende sul verde giardino \\ l'estate è qui, con il suo caldo e il suo \\ dolce profumo di fiori e di frutta \\ e il vento che soffia leggero, come un bacio \\ sulla mia fronte, come un raggio di sole}} \\[25pt]

ITA & \textit{\makecell{Quando il sole splende sulle campagne calde, \\ E il cielo è sereno, e il vento è quieto e lieve,\\ L'estate, con i suoi raggi, ci accoglie, \\ E il cuore, con gioia, si sveglia }} \\[4pt]

\midrule

System & Rispondi alla domanda \\ 
Prompt & Se in una strada costruisco una corsia in più riduco il traffico? Motiva la risposta \\

\midrule
Original & \textit{La risposta è no, non necessariamente riduco il traffico. Ecco alcune motivazioni: \newline 1. **Attrazione di traffico**: la creazione di una corsia in più può attirare più veicoli sulla strada, poiché gli automobilisti potrebbero essere più propensi a utilizzare la strada con più corsie. Ciò può aumentare il traffico sulla strada.  \newline 2. **Cambiamenti di comportamento**: la creazione di una corsia in più può cambiare il comportamento degli automobilisti, che potrebbero essere più propensi a utilizzare la strada per raggiungere i loro obiettivi più rapidamente, aumentando il traffico.} [...] \\[30pt] 

ANITA \citep{polignano2024advanced} & \textit{In generale, non è detto che creando una corsia in più in una strada riducano il traffico.  Ecco alcune possibili motivazioni: \newline 1. **Attrazione di nuovo traffico**: la creazione di una corsia in più può attirare nuovi veicoli sulla strada, poiché la capacità di attraversamento aumenta, il che potrebbe annullare l'effetto di riduzione del traffico previsto.}  [...] \\[30pt]

ITA-full & \textit{La risposta è no, non necessariamente. La costruzione di una corsia in più in una strada non è sufficiente per ridurre il traffico in modo significativo. Ecco alcune ragioni per cui: \newline 1. La corsia in più non è necessariamente una misura per ridurre il traffico. La costruzione di una corsia in più può attirare più traffico nella zona (domanda indotta), poiché gli automobilisti coglieranno l'opportunità di poter usare la nuova corsia} [...] \\[30pt]

ITA & \textit{La risposta è no, non necessariamente riduco il traffico.\newline Costruire una corsia in più in una strada può avere diversi effetti sul traffico, a seconda del contesto e delle condizioni. Ecco alcuni possibili effetti: \newline 1. **Riduzione del traffico**: se la corsia in più è destinata ai mezzi pubblici o ai pedoni e ciclisti, può ridurre il traffico generale sulla strada, } [...] \\[30pt]

\midrule

System & Rispondi alle domande dell'utente \\ 
Prompt & Riassumi il seguente testo: "Il Monte Bianco (Mont Blanc in francese e in arpitano), con un'altitudine di 4805,59 m s.l.m., è la montagna più alta delle Alpi, d'Italia, di Francia e anche d'Europa, secondo una diffusa convenzione geografica, da cui i soprannomi di tetto d'Europa e di Re delle Alpi, condividendo assieme al monte Elbrus nel Caucaso un posto tra le cosiddette Sette Vette (Seven Summits) del pianeta." \\

\midrule
Original & \textit{Il testo descrive il Monte Bianco, la montagna più alta delle Alpi, d'Italia, di Francia e d'Europa, con un'altitudine di 4805,59 metri. È noto anche come "tetto d'Europa" e "Re delle Alpi". Inoltre, è considerato una delle Sette Vette del pianeta, insieme al monte Elbrus nel Caucaso.} \\[4pt] 

ANITA \citep{polignano2024advanced} & \textit{Il testo descrive il Monte Bianco, la montagna più alta d\'Europa, delle Alpi, d'Italia e di Francia, con un'altitudine di 4805,59 metri, noto anche come "tetto d'Europa" e "Re delle Alpi".} \\[4pt] 

ITA-full & \textit{Il testo descrive il Monte Bianco, la montagna più alta delle Alpi, d'Italia, di Francia e d'Europa, con un'altitudine di 4805,59 m s.l.m. È noto anche come "Tetto d'Europa" e "Re delle Alpi" e fa parte delle cosiddette "Sette Vette" del pianeta, insieme al Monte Elbrus nel Caucaso.} \\[4pt]

ITA & \textit{Il testo descrive il Monte Bianco, la montagna più alta delle Alpi, d'Italia, di Francia e d'Europa, con un'altitudine di 4805,59 metri. Inoltre, menziona i soprannomi della montagna, ovvero "Tetto d'Europa" e "Re delle Alpi", e la sua inclusione tra le "Sette Vette" del pianeta, insieme al monte Caucaso} \\[4pt]

\bottomrule
\end{tabular}
}
\caption{Example generations on random prompts for the Llama~3 - Instruct model in all previously proposed versions. Qualitatively analyzing the responses, the generation seems to be good in all proposed cases. Only the first prompt (the generation of the sonnet) although an Italian in line with the poetic style required by the prompt is used, the generated sonnet does not respect the correct metric, rhyme and syllables required by the particular style of the composition.}
\label{tab:llama-open-end-gen}
\end{table*}

\begin{table*}[h]
\centering
% \scalebox{0.83}{
\begin{tabular}{p{0.1\textwidth}p{0.8\textwidth}}
\toprule
\textbf{Model} & \textbf{Prompt / Generations} \\

\midrule
System & Segui le istruzioni dell'utente \\ 
Prompt & Scrivi la prima quartina di un sonetto sull'estate\\
\midrule
Original & \textit{Nel caldo baleno dell'estate,\newline dove il sole splende con ardore,\newline le rose sbocciano in fiore,\newline e il vento canta la dolce canzone.} \\[4pt] 

ITA-full & \textit{La prima quartina di un sonetto è composta da due versi, di solito otto o sei sillabi, e segue il modello ABBAABBACCACC.} \\[4pt]

ITA & \textit{Nell'aria fresca dell'estate,\newline dove il sole splende con calore,\newline il mio cuore si sente il vento,\newline e il mio spirito si eleva.} \\[4pt]

\bottomrule
\end{tabular}
% }
\caption{Example generations on a random prompt for the Phi~3 - Instruct model in all previously proposed versions. The generation is good for the original and ITA model with a strange behavior for the steered ITA-full model where the model explains what a sonnet is instead of writing the requested sonnet. As in the previous case, despite the choice of an Italian in line with the poetic style, the metrics of the typical sonnet composition are not respected.}
\label{tab:phi-open-end-gen}
\end{table*}

\end{document}